\documentclass[11pt]{article}

\usepackage{acl}

\usepackage{times}
\usepackage{latexsym}
\usepackage[switch]{lineno} 
\usepackage[utf8]{inputenc}
\usepackage{CJKutf8}
\usepackage{floatrow}
\usepackage{multirow}
\usepackage{multicol}
\usepackage{array}
\usepackage{booktabs}
\usepackage{graphicx}
\usepackage{amsmath,amsfonts,amssymb}
\usepackage{graphicx}
\usepackage[ruled,noend]{algorithm2e}

\usepackage{xpinyin}
\usepackage{pifont}
\usepackage{subfigure}

\usepackage{xcolor}

\usepackage[T1]{fontenc}
\usepackage[utf8]{inputenc}

\usepackage{microtype}

\usepackage{url}

\usepackage{footnote}
\makesavenoteenv{table}

\newcommand{\cjksong}[1]{{{\begin{CJK*}{UTF8}{gbsn}#1\end{CJK*}}}}
\newcommand{\model}{AiM}

\title{\model{}: Taking Answers in Mind to Correct \\ Chinese Cloze Tests in Educational Applications}

\author{
Yusen Zhang$^1$\thanks{\; Contribution done during internship at Tencent Cloud Xiaowei. Qingyu Zhou is the corresponding author.},~~Zhongli Li$^2$,~~Qingyu Zhou$^2$,~~Ziyi Liu$^{1*}$,\\\textbf{Chao Li}$^3$,~~\textbf{Mina Ma}$^2$,~~\textbf{Yunbo Cao}$^2$,~~\textbf{Hongzhi Liu}$^1$
 \\
  $^1$Peking University,~~$^2$Tencent Cloud Xiaowei,~~$^3$Xiaomi Group \\
   \texttt{\{yusen-zhang0826@stu,lzymail@stu,liuhz@ss\}.pku.edu.cn} \\
   \texttt{\{neutrali,qingyuzhou,minarma,yunbocao\}@tencent.com} \\
    \texttt{lichao51@xiaomi.com}
}

\begin{document}
\maketitle
\begin{abstract}
To automatically correct handwritten assignments, the traditional approach is to use an OCR model to recognize characters and compare them to answers. The OCR model easily gets confused on recognizing handwritten Chinese characters, and the textual information of the answers is missing during the model inference. However, teachers always have these answers in mind to review and correct assignments.
In this paper, we focus on the Chinese cloze tests correction and propose a multimodal approach\footnote{Our codes and data are available at \url{https://github.com/YusenZhang826/AiM}.} (named \model{}). The encoded representations of answers interact with the visual information of students' handwriting. Instead of predicting `right' or `wrong', we perform the sequence labeling on the answer text to infer which answer character differs from the handwritten content in a fine-grained way.
We take samples of OCR datasets as the positive samples for this task, and develop a negative sample augmentation method to scale up the training data.
Experimental results show that \model{} outperforms OCR-based methods by a large margin. Extensive studies demonstrate the effectiveness of our multimodal approach.

\end{abstract}

\section{Introduction}
\label{intro}

\begin{figure}[t]
\small
\centering
\includegraphics[width=0.9\textwidth]{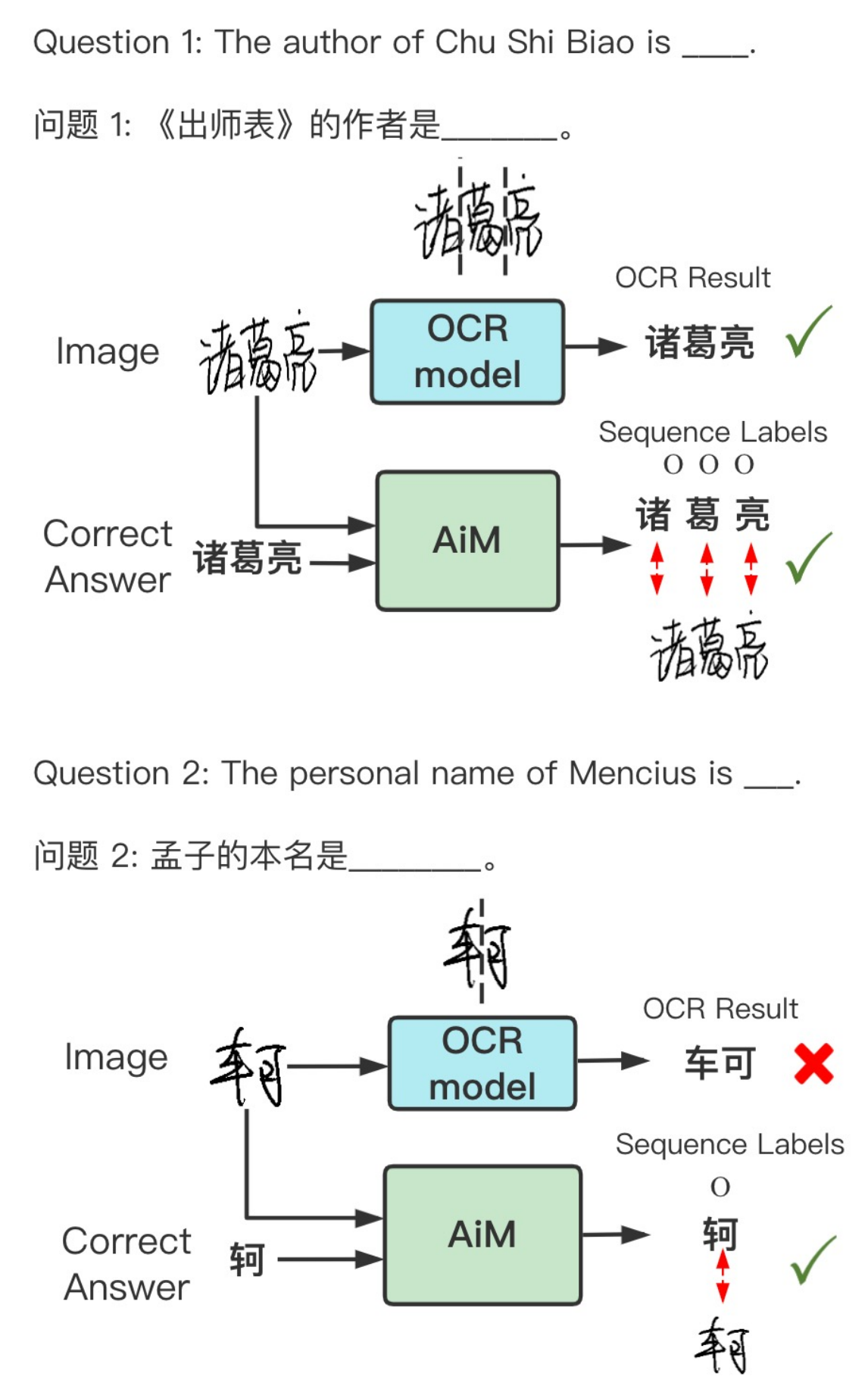}
\caption{\label{fig:Taskdescription}
Task examples and correction methods. The OCR-based method misunderstands student's intent due to the width between ``\cjksong{车}'' and ``\cjksong{可}'' in Question 2. \model{} can tackle this problem since it takes both the handwriting image and the textual answer into consideration.}
\end{figure}

The growing number of students has brought much pressure on teachers to correct assignments manually in the educational field. 
Recently, Optical Characters Recognition (OCR) based methods are widely used in several applications\footnote{Educational applications such as \url{http://kousuan.yuanfudao.com}, \url{https://jiazhang.zuoyebang.com} and \url{https://www.zuoye.ai}.} to automatically complete this task. The OCR model first recognizes the text in the image, and then the OCR output is compared with the correct answer to feedback correction results.
In this pipeline method, the post-comparison mainly relies on the pre-recognition. Without prior knowledge of answers, the OCR model easily gets confused when recognizing handwritten Chinese characters, especially for various handwriting styles, ligatures, and shape-similar characters. 
However, most of them can be distinguished by human teachers, because they take answers in mind at first.

In this paper, we focus on the Chinese cloze correction (CCC) and propose an \textbf{A}nswer-\textbf{i}n-\textbf{M}ind correction model (\model{}) to tackle the above problem.
Figure~\ref{fig:Taskdescription} shows examples of the CCC task and the comparison of the OCR-based method with our method.
We look at Question 2. Obviously, the student knows the correct answer and writes down ``\cjksong{轲}'', but it is recognized as ``\cjksong{车可}'' by the OCR model because of the slightly large width of the hand-written content. Consequently, the correction result is `wrong'.
We hypothesize that the information of correct answers can help the neural model to understand student handwriting. As shown in Figure~\ref{fig:Taskdescription}, \model{} is a multimodal model, which takes the image and the answer text as input. Through the interaction of two modality information, our model understands the handwritten content and feeds back the `right' correct result.

In the \model{} model, the image is encoded to sequential feature representations through Resnet~\cite{he2016resnet}, where each of them represents a fixed-width pixels block. The answer text is encoded by word embedding. 
Then Transformer~\cite{vaswani2017attention} self-attention is adopted to compute contextual representations for each modality. In order to fuse them, we develop a cross-modal attention. It renders the textual representations to interact with the visual information of students' handwriting. 
On the top of \model{}, instead of predicting `right' or `wrong', our model performs sequence labeling on the answer text to infer which character differs from the handwritten content.

To train \model{}, we collect EinkCC, a dataset containing about 5k handwriting images, answers, and correction results of cloze questions, from our educational application. 
Teachers distribute cloze tests in our app, and students practice on the electronic paper hardware.
In addition to EinkCC, OCR datasets can be used for this task. 
We take samples of OCR datasets~\citep{liu2011casia} as the positive samples, and construct negative samples by replacing the label characters with shape-similar ones derived from an open-sourced confusion set~\cite{confusion_set}. The same method also augments EinkCC to scale up the training set.

We pretrain the image encoder in \model{} to get better visual representations, and further train \model{} with the correction objective. Experimental results show that compared with OCR-based methods~\cite{liu2020offline, ppocrv2}, \model{} achieves 11\% accuracy improvements.
Extensive analyses verify that with the interactions between two modalities through our attention mechanism, \model{} can understand student handwriting and ligatures, and more handwritten characters confused by OCR can be predicted well by \model{}.

The main contributions are summarized as follows:
    i) We propose \model{}, a multimodal model for Chinese cloze correction, to make up for shortages of OCR-based methods. 
    ii) We extend OCR datasets using a negative sample augmentation method to fit this task.
    iii) Comprehensive experiments show that \model{} achieves better performance compared with OCR-based methods, and it's effective and necessary to use a multimodal approach to correct Chinese cloze tests.

\begin{figure*}[t]
    \centering
    \includegraphics[width=0.9\textwidth]{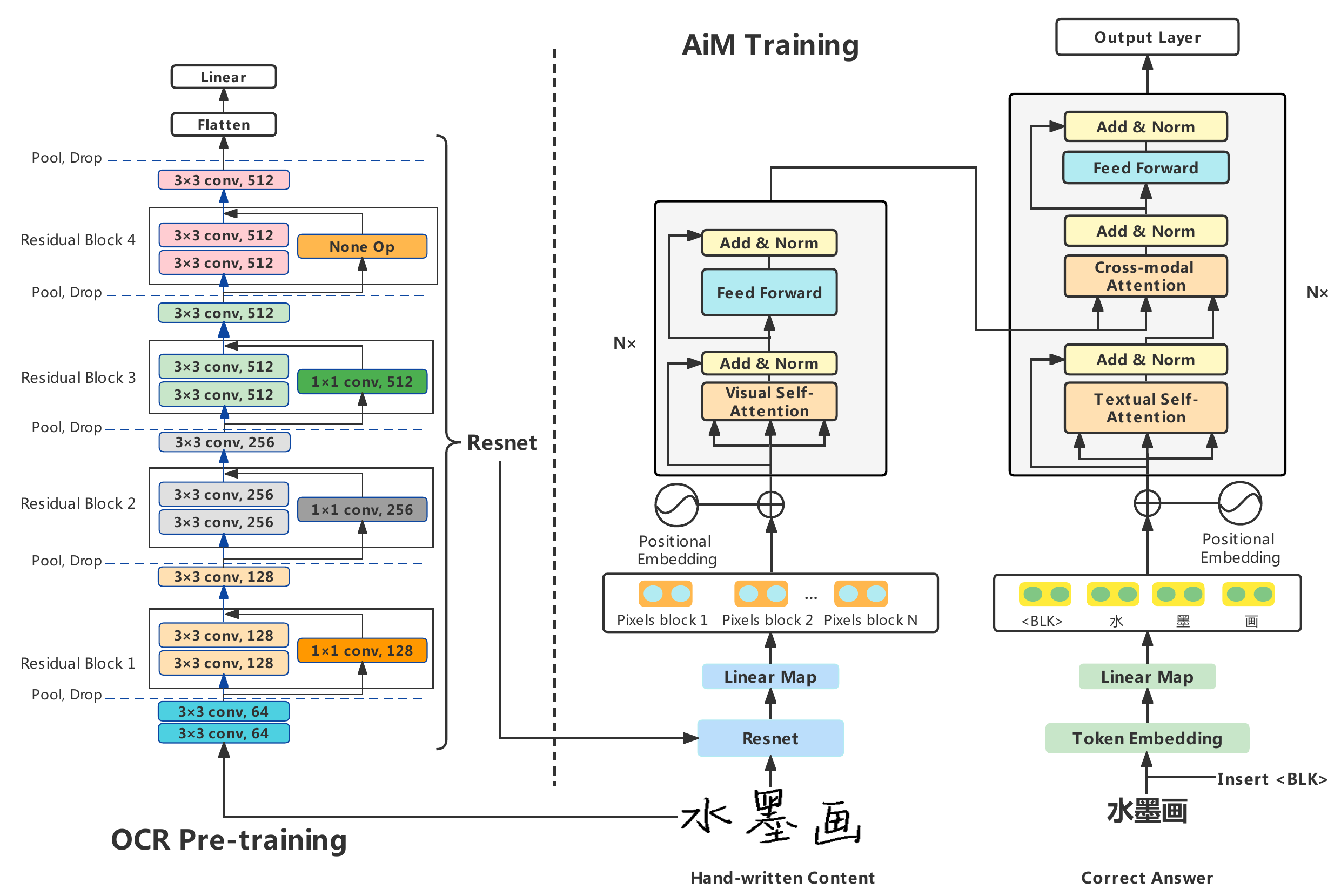}
    \caption{The architecture of \model{}. The image encoder of \model{} is initialized by OCR pretraining. In \model{}, the visual and textual representations are fused by cross-modal attention.}
    \label{fig:MAC_arch}
\end{figure*}

\section{Preliminary}

\subsection{Chinese Cloze Correction}
We first give the description of the Chinese cloze correction (CCC) task in this section. Given a handwriting image $\mathbf{I}$ and the textual answer ${\mathbf{A}=[a_1, a_2, ..., a_m]}$ of the corresponding question, assuming that the handwritten characters in the image is ${\mathbf{C}=[c_1, c_2, ..., c_n]}$, the target of the task is to predict a label $y \in \{0,1\}$. The $y=0$ indicates the correction result is `right' (i.e. the handwritten content is a correct answer, $\mathbf{C}=\mathbf{A}$), otherwise it is `wrong'.

\subsection{Transformer}
Suppose the input of Transformer~\cite{vaswani2017attention} is a pack of embeddings ${\bf X}^{0}=[{\bf x}_1,{\bf x}_2, ...,{\bf x}_{|x|}].$ 
If we have $L$ stacked Transformer blocks, the final output is like:
\begin{equation}
    {\bf X}^{l}=\text{Transformer}_{l}({\bf X}^{l-1}), l\in [1,L]
\end{equation}
where each block consists of a self-attention layer, a feed-forward layer, residual connection~\cite{he2016resnet} and layer normalization.

\paragraph{Self-Attention} For the $l$-th block, the output ${\bf A}_l$ of a self-attention head is:
\begin{equation}
\begin{aligned}
&{\bf Q}={\bf X}^{l-1}{\bf W}^{Q}_{l}, {\bf K}={\bf X}^{l-1}{\bf W}^{K}_{l} \\
& {\bf M}_{i,j}=
\begin{cases}0,&
\text{allow to attend} \\
-\infty,& \text{forbid to attend}
\end{cases}\\
& {\bf A}_l=\text{SelfAttention}({\bf X}^{l-1}) \\
& ~~~~~ =\text{softmax}(\frac{{\bf QK}^{\texttt{T}}}{\sqrt{d_k}}+{\bf M})({\bf X}^{l-1}{\bf W}^{V}_{l})\\
\end{aligned}
\end{equation}
where ${\bf W}^{Q}_{l},{\bf W}^{K}_{l},{\bf W}^{V}_{l}$ can project previous output to queries, keys, and values, respectively. ${\bf M} \in \mathbb{R}^{|x| \times |x|}$ is a mask matrix that controls whether two tokens can attend each other. 

\section{Method}
We first introduce the sequence labeling conversion for the CCC task in Section~\ref{sec:seq_labeling}. Model architecture of \model{} is shown in Section~\ref{sec:arch} and Figure~\ref{fig:MAC_arch}. Finally, we describe the data augmentation and pretraining methods in Section~\ref{sec:data-aug}.

\subsection{Label Space}
\label{sec:seq_labeling}
The CCC task is to feedback `right' or `wrong' on the handwritten content. In this paper, we perform sequence labeling on the answer text, and the corresponding labels are defined as:
\begin{itemize}
    \item \texttt{del}: the current character does not appear in the image.
    \item \texttt{add}: compared to the handwritten content, one or more characters should be inserted between the current and the next character\footnote{A placeholder \texttt{<BLK>} is inserted at the beginning of the textual answers to handle the character missing at the first position.}.
    \item \texttt{sub}: compared to the handwritten content, the current character should be substituted by another one.
\end{itemize}
We get these labels by calculating the edit distance between the answer and the handwritten content\footnote{The handwritten text is annotated or taken from the OCR dataset for calculating the label sequence of \model{}.}.

The BIO annotation~\cite{bio_scheme} is adopted that the label space is \{\texttt{O}, \texttt{B-sub}, \texttt{I-sub}, \texttt{B-del}, \texttt{I-del}, \texttt{B-add}\}. It is similar to the label space of the Grammatical Error Detection/Correction (GED/GEC) task, but our method compares the answer text to the handwritten content, instead of the erroneous sentence to the correct sentence.
After this conversion, \model{} is trained in a fine-grained way. If the predicted sequence only contains `\texttt{O}', the correction result of \model{} is `right', otherwise it is `wrong'.

\subsection{Model Architecture}
\label{sec:arch}
The model architecture of \model{} is shown in Figure \ref{fig:MAC_arch}. Components include: i) an image encoder with Resnet and the self-attention mechanism to extract the visual features, ii) a fusion module with the cross-modal attention to mine the interactions between modalities,  and iii) an output layer to predict the label sequence.

\paragraph{Image Encoder}
To understand the handwritten content in images, we follow~\citet{liu2020offline} to adopt Resnet~\citep{he2016resnet} as the image encoder. The image encoder maps the input image $\mathbf{I}$ to a sequence of visual features $\mathbf{H}_v \in \mathbb{R}^{N_v \times d}$ with a linear transformation:
\begin{equation}
\begin{split}
    \mathbf{H'}_v & =\text{Linear}\left(\text{ResNet}(\bf I)\right) \\
    \mathbf{H}_v  & =\mathbf{H'}_v + \mathbf{P}_v
\end{split}
\end{equation}
where $N_v =\frac{max\_width}{32}$, ${max\_width}$ is the maximum width of input images and $d$ is the hidden size of the visual representation. Each element in sequence represents a fixed-width pixels block in the image. 
Besides, there is a learnable positional embedding matrix $\mathbf{P}_v \in \mathbb{R}^{N_v  \times d}$ where each row is a positional representation for each element in $\mathbf{H}_v$ to capture the location information. 

We perform a padding operation on the image with extra white pixels blocks to ensure all images have the same width. Then we adopt Transformer blocks to capture the contextual information in visual modality and the $l$-th output is:
\begin{equation}
\begin{aligned}
&{\bf S}^{l}_{v}=\text{Transformer}({\bf S}^{l-1}_{v}), l \in [1, L_v].  \\
\end{aligned}
\end{equation}
where $L_v$ is the number of Transformer blocks. Notes that ${\bf S}^{0}_{v}=\mathbf{H}_v$ and $\mathbf{S}_v = \mathbf{S}^{L_v}_{v}$.

\paragraph{Fusion Module}
Assuming that the input answer has $N_t$ characters, the answer characters are first converted to dense vectors $\mathbf{X}=[x_1, x_2, ..., x_{N_t}]$ through a word embedding.
The linear transformation and positional embedding are also employed to compute the textual representation $\mathbf{H}_t \in \mathbb{R}^{N_t \times d}$ as follows:
\begin{equation}
\begin{split}
    \mathbf{H'}_t & =\text{Linear}(\mathbf{X}) \\
    \mathbf{H}_t  & =\mathbf{H'}_t + \mathbf{P}_t
\end{split}
\end{equation}
where $\mathbf{P}_t$ is the positional embedding matrix.

Then our fusion block further encodes and merges the information of textual and visual modality.
As shown in Figure~\ref{fig:MAC_arch}, each fusion block contains a textual self-attention layer and a cross-modal attention layer.  
Self-attention mechanism is employed to encode textual representations:
\begin{equation}
\begin{aligned}
&{\bf S}'_{t}=\text{SelfAttention}({\mathbf{H}_t}) \\
&{\bf S}_{t}=\text{LayerNorm}(\mathbf{H}_t+{\bf S}'_{t})
\end{aligned}
\end{equation}
To capture the interactions between them, we develop a cross-modal attention as follows:
\begin{equation}
\begin{aligned}
&{\bf Q}=\mathbf{S}_t{\bf W}^{Q}, {\bf K}=\mathbf{S}_v{\bf W}^{K} \\
&{\bf M}^{f}_{i,j}=
    \begin{cases}-\infty,&
    \text{padding token or pixels block} \\
    0,& \text{otherwise}
    \end{cases}\\
&{\bf S}_{f}=\text{softmax}(\frac{{\bf QK}^{\texttt{T}}}{\sqrt{d}}+{\bf M}^{f})({\bf S}_{v}{\bf W}^{V}) \\
\end{aligned}
\end{equation}
where $\mathbf{W}^Q$ projects $\mathbf{S}_t$ to queries, $\mathbf{W}^K$ and $\mathbf{W}^V$ project $\mathbf{S}_v$ to keys and values, and ${\bf M}^{f} \in \mathbb{R}^{N_t \times N_v}$ is the mask matrix to ensure only valid tokens and pixels blocks can attend to each other. The fused representations $\mathbf{S}_f$ are followed by the feed-forward layer, residual connection, and layer normalization.

Finally, we stack our fusion blocks to obtain more informative fused features.
It can be seen that our fusion module is similar to the Transformer decoder. Different from the cross-attention of Transformer, our cross-modal attention merges the visual and textual information, which allows each valid answer character to attend to all valid pixels blocks, without the order restriction of language modeling. Besides, the Transformer decoder decodes words one by one, but our module outputs fused features at once.



\paragraph{Output Layer} We denote the above fused features as $\mathbf{H}_f$. To project the hidden representation to the space of labels, the fusion features is fed to the fully-connected layer and a softmax function to get the final output ${{\bf O} \in \mathbb{R}^{N_t\times m}}$:
\begin{equation}
    {\bf O}=\text{softmax}(\mathbf{H}_f{\bf W}_o)
\end{equation}
where ${\bf W_o} \in \mathbb{R}^{d \times m}$ is the weight matrix and $m$ is the number of labels. During the training, we apply the cross-entropy function as our correction objective and the training loss is computed as:
\begin{equation}
    \mathcal{L} = -\sum_{i=1}^{N_t}~\text{log}~p(i,k_i),~~ p(i,k_i) \in \mathbf{O}
\end{equation}
where $k_i$ is the label of the $i$-th character and $p(i,k_i)$ is the probability of the $i$-th character being predicted to label $k_i$.

\subsection{Data Augmentation and Pretraining}
\label{sec:data-aug}

\begin{algorithm}[t]
\small
\SetAlgoNoLine
\DontPrintSemicolon
\KwIn{$\mathcal{S}=\{({\bf I}_i, {\bf C}_i, {\bf A}_i, {\bf L}_i, y_{i})\}_{i=1}^{N}$, where ${\bf I}_i$ is the image, ${\bf C}_i$ is the handwritten content, ${\bf A}_i$ is the answer text, ${\bf L}_i$ is the label sequence of \model{} and $y_i$ is the correction result.}
$D \gets \mathcal{S}$\;
\For{$({\bf I}_i, {\bf C}_i, {\bf A}_i, {\bf L}_i, y_{i})$ \textbf{in} $\mathcal{S}$}{
    \Repeat{\text{Random times}}{
        $\hat{{\bf A}}_i \gets {\bf A}_i$\;
        Randomly select an index $j$ in $\hat{{\bf A}}_i$\;
        $\hat{{\bf A}}_i[j] \gets $ ${\bf A}_i[j]$ 's shape-similar character\;
        Get $\hat{{\bf L}}_i$ comparing ${\bf C}_i$ and  $\hat{{\bf A}}_i$\;
        $\hat{y}_i=1$\;
        $D$.append ($({\bf I}_i, {\bf C}_i, \hat{{\bf A}}_i, \hat{{\bf L}}_i, \hat{y}_i)$)\;
    }
    \Repeat{\text{Random times}}{
        $\hat{{\bf A}}_i \gets {\bf A}_i$\;
        Randomly select an index $j$ in $\hat{{\bf A}}_i$\;
        Delete $\hat{{\bf A}}_i[j]$ \;
        Get $\hat{{\bf L}}_i$ comparing ${\bf C}_i$ and  $\hat{{\bf A}}_i$\;
        $\hat{y}_i=1$\;
        $D$.append ($({\bf I}_i, {\bf C}_i, \hat{{\bf A}}_i, \hat{{\bf L}}_i, \hat{y}_i)$)\;
        }
    
    \Repeat{\text{Random times}}{
        $\hat{{\bf A}}_i \gets {\bf A}_i$\;
        Randomly select an index $j$ in $\hat{{\bf A}}_i$\;
        Insert a common Chinese character at $\hat{{\bf A}}_i[j]$ \;
        
        Get $\hat{{\bf L}}_i$ comparing ${\bf C}_i$ and  $\hat{{\bf A}}_i$\;
        $\hat{y}_i=1$\;
        $D$.append ($({\bf I}_i, {\bf C}_i, \hat{{\bf A}}_i, \hat{{\bf L}}_i, \hat{y}_i)$)\;
    }
}
\KwOut{the augmented dataset $D$}
\caption{Negative Sample Augmentation}
\label{alg-data-aug}
\end{algorithm}

\paragraph{Data Augmentation} 
The sample of the OCR dataset contains an image and the corresponding written text, which can be easily extended for the CCC task.
We directly take the annotated text as the answer to augment our positive CCC samples.
Then we develop a negative sample augmentation method. Given a CCC sample, we keep the image unchanged and modify the answer text. The modifications include random character insertion, deletion and substitution. Especially for the character substitution, we attempt to construct hard negative samples by replacing the original character with shape-similar ones. The shape-similar characters are derived from an open-sourced confusion set\footnote{It is taken from \href{https://github.com/ACL2020SpellGCN/SpellGCN/blob/master/data/gcn_graph.ty_xj/SimilarShape_simplied.txt}{URL}.}~\citep{confusion_set}.
The pseudo code of negative sample augmentation is presented in Algorithm \ref{alg-data-aug}.

\paragraph{Pretraining} Before the \model{} training, we first pretrain our image encoder with the OCR objective. We follow the \citet{liu2020offline} to use CTC~\citep{graves2006connectionist} as the loss function and apply high dropout rates after each max-pooling layer. As shown on the left side in Figure~\ref{fig:MAC_arch}, the input image is converted to a sequence of dense representations through Resnet. A linear layer is added on the top of Resnet to transform the space dimension to the size of the vocabulary.

\begin{table}[t]
\small
    \centering
    \caption{Data statistics. ``\#img'' means the number of handwriting images. ``\#sample'' means the number of CCC samples. In each train or dev set, the ``\#sample'' is larger than ``\#img'' because of our data augmentation in Section~\ref{sec:data-aug}.}
    \label{tab:OCR partition}
    \begin{tabular}{@{}p{0.9cm}|p{0.65cm}p{0.85cm}|p{0.45cm}p{0.85cm}|p{0.35cm}p{1.0cm}@{}}
    \toprule
    \multirow{2}{*}{Dataset} & \multicolumn{2}{c|}{Train set} & \multicolumn{2}{c|}{Dev set} & \multicolumn{2}{c}{Test set}\\
    & \#img & \#sample & \#img & \#sample & \#img & \#sample \\
    \midrule
    EinkCC & 4256 & 10610 & - & - & 673 & 673 \\
    HWCC  & 41781 & 166984 & 10499 & 41767 & - & - \\
    SynCC & 150594 & 602376 & - & - & - & - \\
    \midrule
    Total & 196631 & 779970 & 10499 & 41767 & 673 & 673 \\
    \bottomrule 
    \end{tabular}
\end{table}

\section{Experiment}
In this section, we first introduce datasets, baselines, and other details of our experiments. Then we show experimental results and perform analyses from different views.
\subsection{Dataset}

\paragraph{EinkCC} 
EinkCC is collected from our educational application. 
Teachers distribute cloze tests in our app, and students practice on the e-ink display hardware. When students finish tests, teachers can correct them and feedback to students.
Each sample in EinkCC contains a student's handwriting image, the answer text and the correction result marked by teachers. It mainly covers the dictation of ancient poetry, the idiom application, and the reading comprehension. Besides, we manually annotate the image text for the OCR training and the \model{} label sequence calculation.

\paragraph{HWCC} CASIA-HWDB~\cite{liu2011casia} is a benchmark for the handwritten Chinese text recognition(HCTR) task. We select the HWDB 2.x set to be the positive CCC samples.

\paragraph{SynCC} 
To further enlarge the scale of CCC datasets, we first build a synthetic OCR dataset. The details of the data construction are presented in Appendix.
Then all OCR samples are taken as the positive CCC samples.

We split EinkCC into train and test sets, and split HWCC into train and dev sets. All train sets and dev sets are extended by our negative sample augmentation method of Section~\ref{sec:data-aug}. The data statistics are shown in Table~\ref{tab:OCR partition}.



\begin{table}[t]
\small
    \centering
    \begin{tabular}{llc}
    \toprule
    Dataset & OCR Model & CER\\
    \midrule
    \multirow{3}{*}{HWCC} & {PP-OCRv2} & 0.3936 \\
    & {CNN-CTC-CBS} & 0.2943 \\
    & {Resnet-CTC} & \textbf{0.0663} \\
    
    \midrule
    \multirow{3}{*}{EinkCC} & {PP-OCRv2} &  0.3115 \\
    & {CNN-CTC-CBS} & 0.2325\\
    & {Resnet-CTC} & \textbf{0.1661}\\
    \bottomrule 
    \end{tabular}
    \caption{The OCR performance on dev and test sets.}
    \label{tab:baseline_ocr}
\end{table}

\begin{table*}[t]
\small
    \centering
    \begin{tabular}{l l ccc cccc }
    \toprule
       \multirow{2}{*}{Dataset} & \multirow{2}{*}{Model} &\multicolumn{3}{c}{Sequence level} & \multicolumn{4}{c}{Binary level} \\
    &  & P &  R &  F1 & P & R & F1 & Acc\\ 
    \midrule
   \multirow{5}{*}{HWCC} 
   & PP-OCRv2 & 0.3590 & 0.4661 & 0.4057 & 0.7508 & \textbf{0.9997} & 0.8576 & 0.7510 \\ 
   & CNN-CTC-CBS & 0.4584 & 0.5563 & 0.5026 & 0.7604 & 0.9994 & 0.8637 & 0.7635 \\ 
   & Resnet-CTC  & 0.8157 & 0.8877 & 0.8502 & 0.8578 & 0.9989 &0.9230 & 0.8751 \\ 
   & \model{}$_{\text{wo-PT}}$ & 0.4893 & 0.5602 & 0.5223 & 0.9083 & 0.9306 & 0.9193 & 0.8571 \\  
   & \model{} & \textbf{0.9735} & \textbf{0.9717} & \textbf{0.9726} & \textbf{0.9926} & 0.9937 & \textbf{0.9932} & \textbf{0.9898} \\   
  \midrule
    \multirow{5}{*}{EinkCC} 
   & PP-OCRv2 & 0.1146 & 0.6847 & 0.1964 & 0.2128 & \textbf{1.0000} & 0.3509 & 0.4502 \\ 
   & CNN-CTC-CBS & 0.1829 & \textbf{0.8108} & 0.2985 & 0.2710 & \textbf{1.0000} & 0.4264 & 0.6003 \\ 
   & Resnet-CTC & 0.2311 & 0.7748 & 0.3561 & 0.3322 & 0.9964 & 0.4987 & 0.7013 \\ 
   & \model{}$_{\text{wo-PT}}$ &0.0568 & 0.2252 & 0.0907 & 0.2025 & 0.6497 & 0.3088 & 0.5676 \\ 
   & \model{} & \textbf{0.2795} & 0.5766 & \textbf{0.3765} & \textbf{0.4262} & 0.7799 & \textbf{0.5512} & \textbf{0.8113} \\ 
  \bottomrule
    \end{tabular}
    \caption{Main results on dev and test sets. {\model{}}$_{\text{wo-PT}}$ is the \model{} model without the OCR pretraining. Resnet-CTC is trained using the same  data source as \model{}, which is a strong OCR baseline in a fair comparison.}
    \label{tab:overall-performance}
\end{table*}

\begin{table}[t]
\small
    \centering
    \begin{tabular}{ p{3.0cm} c }
    \toprule
       Model & CER \\
  \midrule
    { Resnet-CTC}$_{\text{wo-Syn}}$ & 0.1916 \\
    { Resnet-CTC}$_{\text{wo-E}}$ & 0.2995  \\
    { Resnet-CTC}$_{\text{wo-H}}$ &  0.3304 \\
   { Resnet-CTC} & \bf{0.1661} \\
  \bottomrule
    \end{tabular}
\caption{Data ablation of the Resnet-CTC OCR performance on the EinkCC test set.}
\label{tab:ocr-ablation}
\end{table}

\subsection{Evaluation Metrics}
We use the character error rate (CER) to evaluate the OCR performance in the pretraining stage, and compute the following metrics to evaluate the CCC performance:
\paragraph{Sequence level} We use the widely-used metrics, Precision ($P$), Recall ($R$), and $F1$, to measure the quality of label sequences outputted by \model{}.
\paragraph{Binary level} We transform label sequences to binary labels (`right' or `wrong') to compute the accuracy of binary classification. We report the $P,R, \text{and } F1$ of negative samples (i.e. $y=1$) to evaluate model's ability to identify wrong answers.

\subsection{Implementation Details and Baselines}
Our training dataset $\{({\bf I}_i, {\bf C}_i, {\bf A}_i, {\bf L}_i, y_{i})\}_{i=1}^{N}$ contains the image ${\bf I}_i$, the handwritten content ${\bf C}_i$, the answer text ${\bf A}_i$, the label sequence of \model{} ${\bf L}_i$ and the correction result $y_i$.
We pretrain the image encoder Resnet on the subset $\{({\bf I}_i, {\bf C}_i)\}_{i=1}^{N}$, and train the \model{} model on the subset $\{({\bf I}_i, {\bf A}_i, {\bf L}_i)\}_{i=1}^{N}$.

In the OCR pretraining, we follow \citet{liu2020offline} to set the number of convolution blocks to \{2,3,1,4\}. The height of every input image is set to 128 pixels and the width is scaled to the corresponding value, so the shape of each pixels block is $128\times32$. The learning rate is set to 1e-3, the batch size is 8 and the training epoch is 15.

In the \model{} training, the learning rate is 1e-4, the batch size is 8 and the training epoch is set to 14. The dimension of hidden states $d$ is 768. The number of image encoder blocks $N_{\text{enc}}$ and fusion module blocks $N_{\text{fus}}$ in \model{} are both 2.
We adopt AdamW optimizer~\cite{loshchilov2017adamw} and cosine-annealing strategy. To accelerate training, the parameters in Resnet are frozen when training \model{}. We ignore punctuation in the text.

OCR-based methods are our baselines, where the correction results are derived from the post-comparison between the OCR results and answers. The following OCR models are evaluated: 
{\bf Resnet -CTC} is the image encoder of \model{} (Resnet) trained with the CTC function on the subset $\{({\bf I}_i, {\bf C}_i)\}_{i=1}^{N}$.
{\bf PP-OCRv2}~\citep{ppocrv2} is an open-sourced widely-used OCR model in Chinese recognition, and we take its text recognition server model\footnote{\url{https://github.com/PaddlePaddle/PaddleOCR}.} for evaluation.
{\bf CNN-CTC-CBS}~\citep{liu2020offline} is a handwritten text-line recognition model, and we take the released well-trained model\footnote{\url{https://github.com/intel/handwritten-chinese-ocr-samples}.} for evaluation.
Table~\ref{tab:baseline_ocr} shows the OCR performance of baselines on our datasets.




\begin{table}[t]
\small
    \centering
    \begin{tabular}{l ccc}
    \toprule 
    \multirow{2}{*}{Model}& \multicolumn{3}{c}{Sequence Level}\\
    & P & R & F1 \\
    \midrule
    { \model{}}$_{\text{wo-Syn}}$& 0.0212 & 0.1712 & 0.0378 \\

    { \model{}}$_{\text{wo-E}}$& 0.1857 & 0.3964 & 0.2529 \\

    { \model{}}$_{\text{wo-H}}$& 0.2660 & 0.4865& 0.3440 \\

    { \model{}} & \bf{0.2795} & \bf{0.5766} & \bf{0.3765} \\
    
    \bottomrule 
    \end{tabular}
\caption{Data ablation of the \model{} performance at the sequence level on the EinkCC test set.}
\label{tab:mac-ablation}
\end{table}

\begin{figure}[t]
\centering
\subfigure[Resnet performance at binary level]{
\label{fig:resnet-ab} 
\includegraphics[width=0.9\textwidth]{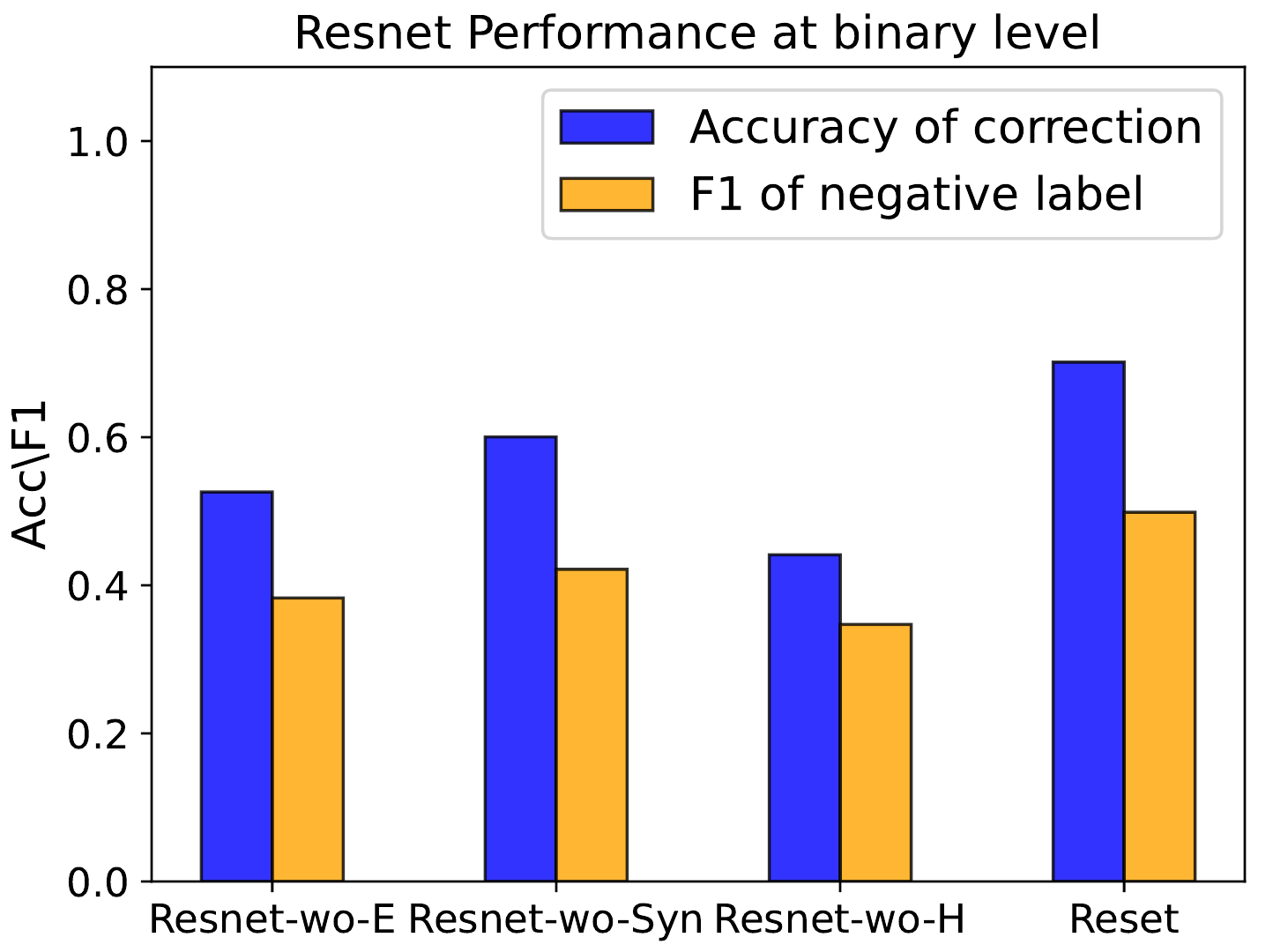}}
\hspace{0.0in}
\subfigure[\model{} performance at binary level]{
\label{fig:mac-b}
\includegraphics[width=0.9\textwidth]{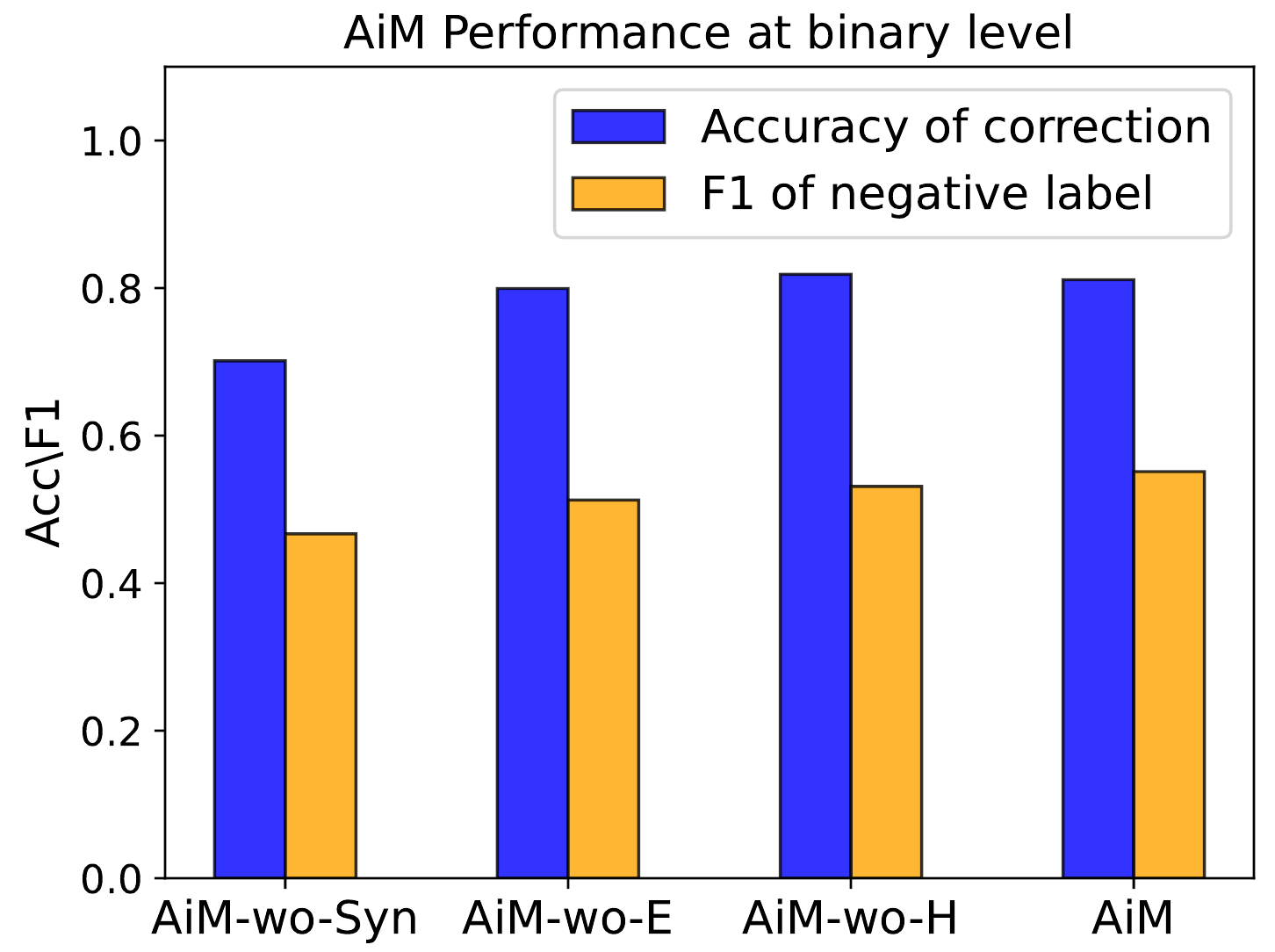}}
\caption{Data ablation of the \model{} and its baseline performance at the binary level on the EinkCC test set.}
\label{fig:data-ab} 
\end{figure}

\subsection{Results and Analyses}
The main results are shown in Table~\ref{tab:overall-performance}. We evaluate OCR models at the OCR level and the binary level. For \model{} and its variants, we evaluate them at the sequence level and the binary level. We describe our observations from the following perspectives:
\begin{itemize}
    \item How do OCR models perform in CCC?
    \item Does \model{} improve the performance and whether the two modalities are well fused?
    \item Do data augmentation and pretraining work and what's the impact of data source?
\end{itemize}

\subsubsection{Limitation of OCR-based Method}
As shown on Table~\ref{tab:overall-performance}, the recall of OCR-based methods is extremely high (nearly 1.0), but the precision and F1 are much lower. This means that OCR models can correct almost all wrong handwritten answers but mark many students' right text as wrong.
Oppositely, \model{} improves the precision, F1, and accuracy with large margins. 

\subsubsection{Influence of Data Source}
We conduct an ablation study on the training data. The suffix `-wo-E', `-wo-Syn', `-wo-H' means the model is trained without the EinkCC training set, SynCC set and HWCC training set, respectively. The performance of Resnet and its variants is shown in Table~\ref{tab:ocr-ablation} and Figure~\ref{fig:resnet-ab}. The CER reaches the lowest level when Resnet is trained with all training sets, which can prove the necessity of data extension for OCR models. Table \ref{tab:mac-ablation} and Figure \ref{fig:mac-b} show the performance of \model{}. Removing any training set, all metrics at sequence level drops, as well as the binary level. It means sufficient data is valuable in the CCC task. Meanwhile, comparing Figure \ref{fig:resnet-ab} and \ref{fig:mac-b}, although the data scale decreases, the performance of \model{} is always better than Resnet, which demonstrates the \model{} is more robust and stable to data variation. 

Notice that the model performance are always better on the HWCC dev set than it on the EinkCC test set. This is because all negative samples of HWCC are synthetic samples that are augmented in the same way. This suggests that AiM learns the error patterns well from our negative sample augmentation.


\cjksong{
\begin{table}[t]
    \centering
    \small
    \begin{tabular}{l c c}
    \toprule
         Image:
        & \includegraphics[width=20mm, height=4mm]{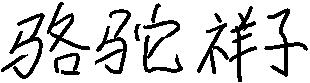}
      &\includegraphics[width=15mm, height=4mm]{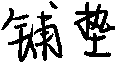}\\
          Handwritten:  & 骆驼祥子& 铺垫\\
          Answer:  & 骆驼祥子 & 铺垫\\
         \hline
          OCR result: & 骆{\bf\underline{马它}}祥子& 铺\bf{\underline{挚}}\\
          \model{} output labels: &O O O O O & O O O\\
         \hline
          Correction label:  & \checkmark & \checkmark\\
          Correcting by OCR: &\ding{53} &\ding{53} \\
          Correcting by \model{}: &\checkmark& \checkmark\\
    \bottomrule
    \end{tabular}
    \caption{Examples on EinkCC test set. Resnet gives the wrong outputs in all examples. The mistakes made by Resnet are shown in bold with underline, while \model{} predicts correctly. Notes that the first label is for the placeholder `<BLK>'.}
    \label{tab:\model{}vsOCR}
\end{table}
}

\begin{table*}[t]
    \centering
    \small
    \begin{tabular}{c c c ccc cccc} 
    \toprule
         ~&~&~&\multicolumn{3}{c}{Sequence level} & \multicolumn{4}{c}{Binary level} \\
         \hline
         $N_{\text{enc}}$ & $N_{\text{fus}}$&Text Self-Att in Fusion & P & R & F1 &P & R & F1& Acc\\
         \hline
         1 & 1 & \ding{53} & 0.1150 & 0.3513 & 0.1733 & 0.3515 & 0.7099 & 0.4702 & 0.7623 \\
         1 & 1 & \checkmark & 0.1773 &  0.4775 & 0.2585 & 0.3794 & \bf{0.7908} & 0.5128 & 0.7727 \\
         2 & 2 & \ding{53} & 0.1235 & 0.3694 & 0.1851 & 0.3630 &  0.7161 & 0.4818 & 0.7667\\
         2 & 2 & \checkmark & \bf{0.2795} & \bf{0.5766} & \bf{0.3765} & \bf{0.4262} & 0.7799& \bf{0.5512} & \bf{0.8113} \\
         3 & 3 & \checkmark & 0.2501& 0.5495 & 0.3436 & 0.3682 &  0.7398 & 0.4917 & 0.7727\\
    \bottomrule
    \end{tabular}
    \caption{An ablation study of components in \model{} on EinkCC test set.}
    \label{tab:ablation}
\end{table*}

\begin{figure*}[t]
    \centering
    \subfigure[]{\includegraphics[width=0.48\textwidth]{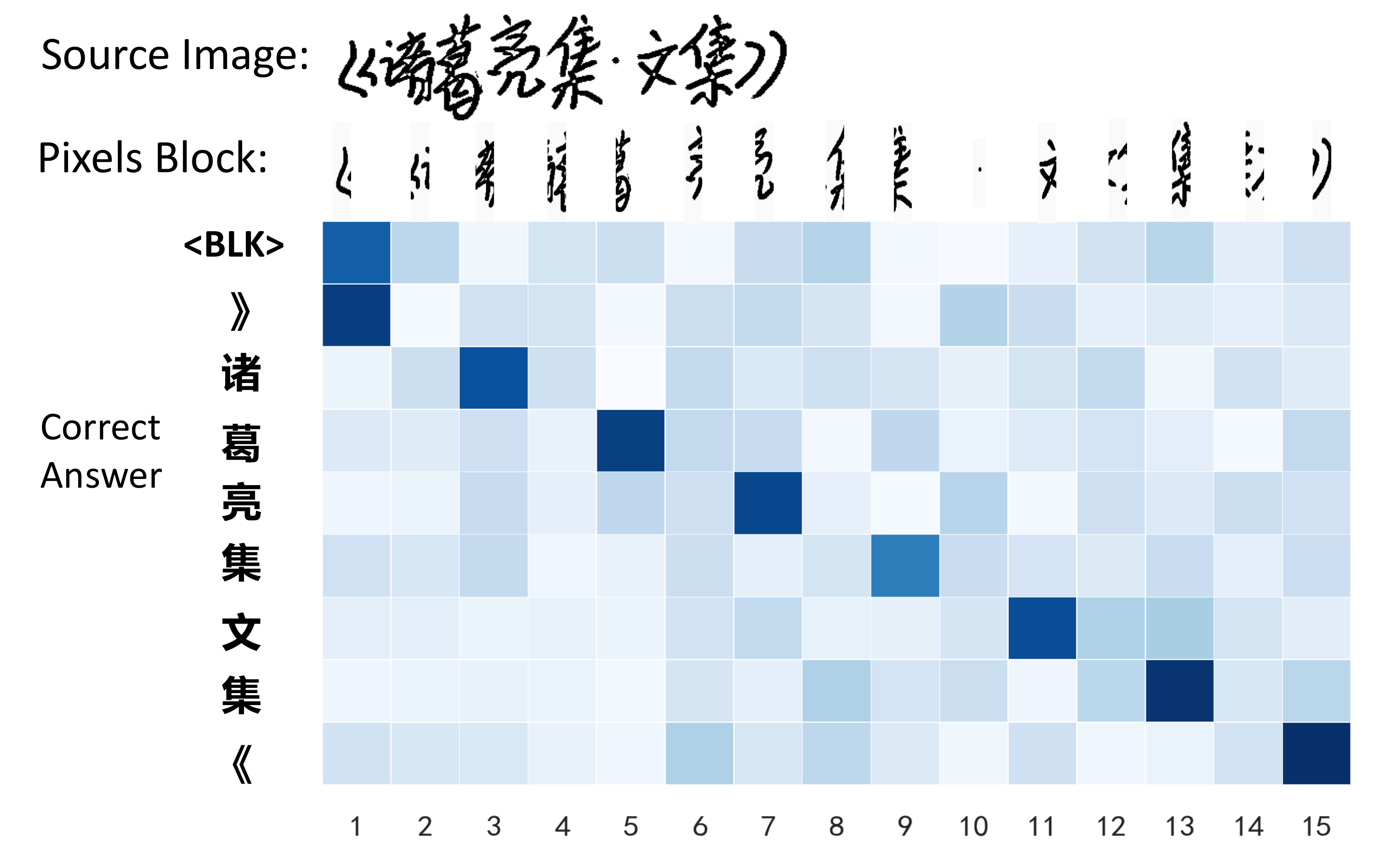}}
    \subfigure[]{\includegraphics[width=0.48\textwidth]{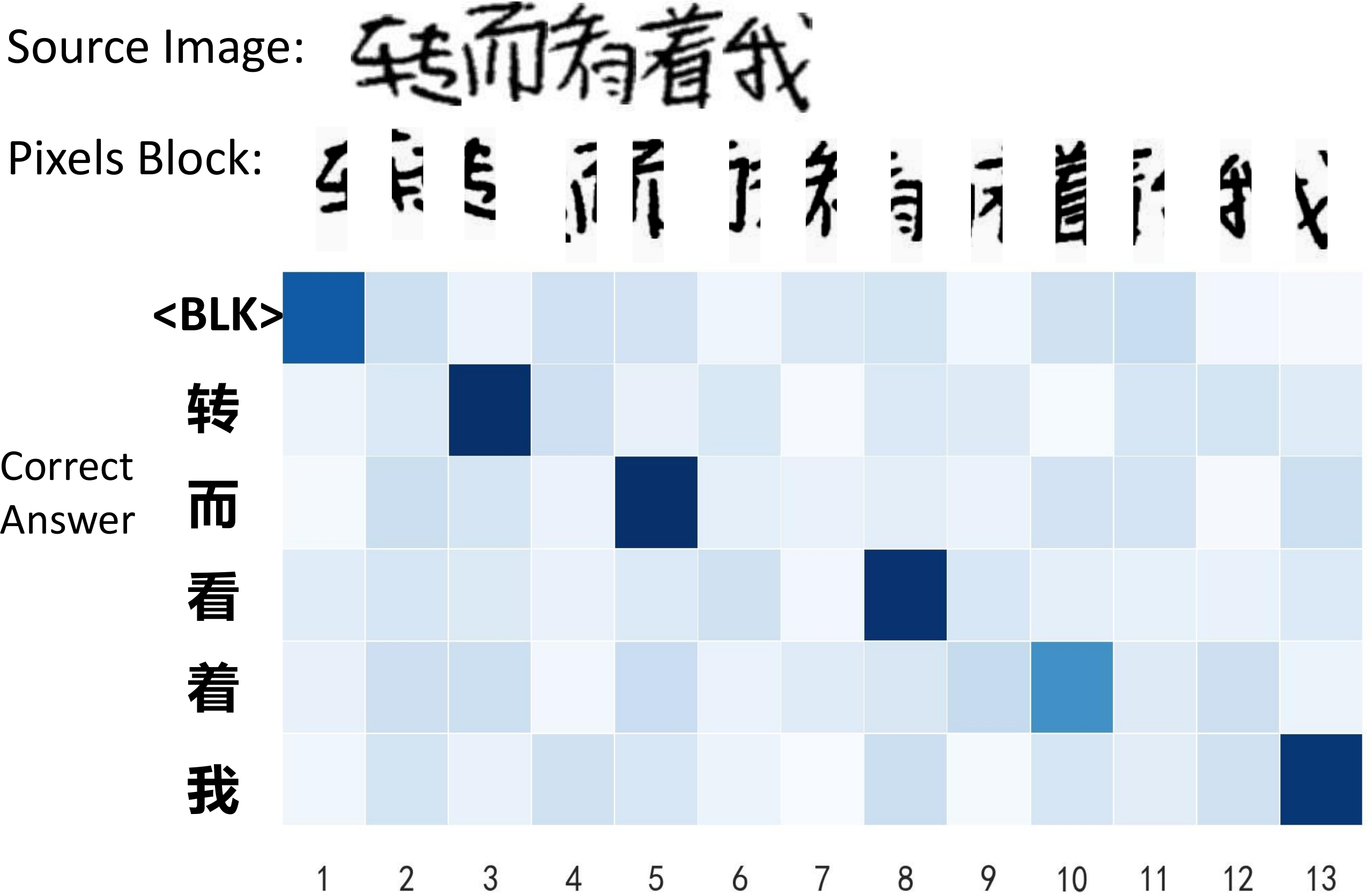}}
    \caption{Cross-modal attention visualization. The figure shows the cross-attention scores between characters in answer and the pixels blocks in image. } 
    \label{fig:attenion-visual}
\end{figure*}

\subsubsection{Effectiveness of Multimodal}
As shown in Table~\ref{tab:overall-performance}, with \model{}, the recall of negative label decreases\footnote{We perform an analysis on the recall drop in Appendix.} but the precision and F1-score increase significantly at the binary level. We give two examples on EinkCC in Table \ref{tab:\model{}vsOCR} to further analyze the effectiveness of \model{}. Obviously, Resnet generates wrong outputs while \model{} makes no mistake. In the first example, Resnet considers one left-right structured character as two independent characters, which never happens when manually correcting. In the second example, Resnet gets confused by the shape-similar characters.  \model{} predicts successfully in these examples. 
More predictions on the dev and test sets are presented in Appendix to illustrate that \model{} is able to identify all situations of the mismatch between the image and answer text. 

Comparing \model{} and {\model{}}$_{\text{wo-PT}}$ in Table~\ref{tab:overall-performance}, without the pretraining, the performance of \model{} drops significantly, even worse than Resnet. It demonstrates that visual modality features can be better understood through pretraining. 

\subsubsection{Learning of Modal Fusion}
 We conduct experiments to explore the contribution of each component. Experimental results are shown in Table~\ref{tab:ablation}. \model{} reaches a better performance when encoding text information with the self-attention than only using token embeddings. This suggests that self-attention brings better textual representation. Meanwhile, stacking each module with a proper number of layers helps \model{} capture interactions between modalities.
 We visualize the cross-modal attention scores in Figure~\ref{fig:attenion-visual}. We find that characters manage to attend the most relative pixels block. 
 For instance, in this sub-figure (a), the character ``\cjksong{集}'' appears twice in different positions, and the model captures the context feature and both of them attend the corresponding pixels block in order.
 It shows that cross-modal attention can match multimodal information effectively.

\section{Related Work}

\subsection{\bf OCR Model}
OCR models have evolved for a long period. Researchers usually used hybrid CNN and RNN architectures~\cite{breuel2017high} with CTC~(Connectionist temporal classification) loss~\cite{graves2006connectionist}. 
For handwritten Chinese text recognition (HCTR) task, \citet{liu2020offline} has achieved the state-of-the-art performance. They use the simple end-to-end CNN-CTC method and ease the overfitting problem with a high-rate dropout strategy. 

\subsection{\bf Multimodal Model}
Multimodal models have attracted more and more attention and have been applied in many fields, such as visual question answering~\citep{vqa}, Chinese spell checking~\citep{csc,li-etal-2022-past} and other applications~\citep{toto2021audibert,hu2021detection,aguilar2019mmemotion}.
Unsupervised pretraining~\citep{devlin2018bert} has provided informative representations and fine-tuning techniques~\citep{cui-etal-2019-fine,li-etal-2021-improving-bert} have brought further performance gains.
Several large-scale pre-trained models are utilized in multimodal models~\cite{anderson2018bottom,toto2021audibert,csc}. To capture multimodal features, one common method is to concatenate encoders' output from two sides and then feed to the downstream multi-layer perceptrons~\cite{nie2021mlp}. While some works use attention mechanism to get fusion representation~\cite{lu2019vilbert,tan2019lxmert,tsai2019mmtransformer}.

\section{Conclusion}
In this paper, we propose a multimodal model \model{} to effectively correct Chinese cloze tests. \model{} employs cross-modal attention to understand the correlation between modalities. We collect data from different sources and develop a data augmentation method. Experiments show that \model{} outperforms traditional OCR-based methods with over 11\% accuracy improvements.

\bibliographystyle{acl_natbib}
\bibliography{coling2020}

\begin{thebibliography}{24}
\expandafter\ifx\csname natexlab\endcsname\relax\def\natexlab#1{#1}\fi

\bibitem[{Aguilar et~al.(2019)Aguilar, Rozgi{\'c}, Wang, and
  Wang}]{aguilar2019mmemotion}
Gustavo Aguilar, Viktor Rozgi{\'c}, Weiran Wang, and Chao Wang. 2019.
\newblock Multimodal and multi-view models for emotion recognition.
\newblock \emph{arXiv preprint arXiv:1906.10198}.

\bibitem[{Anderson et~al.(2018)Anderson, He, Buehler, Teney, Johnson, Gould,
  and Zhang}]{anderson2018bottom}
Peter Anderson, Xiaodong He, Chris Buehler, Damien Teney, Mark Johnson, Stephen
  Gould, and Lei Zhang. 2018.
\newblock Bottom-up and top-down attention for image captioning and visual
  question answering.
\newblock In \emph{Proceedings of the IEEE conference on computer vision and
  pattern recognition}, pages 6077--6086.

\bibitem[{Antol et~al.(2015)Antol, Agrawal, Lu, Mitchell, Batra, Zitnick, and
  Parikh}]{vqa}
Stanislaw Antol, Aishwarya Agrawal, Jiasen Lu, Margaret Mitchell, Dhruv Batra,
  C.~Lawrence Zitnick, and Devi Parikh. 2015.
\newblock \href {http://arxiv.org/abs/1505.00468} {{VQA:} visual question
  answering}.
\newblock \emph{CoRR}, abs/1505.00468.

\bibitem[{Breuel(2017)}]{breuel2017high}
Thomas~M Breuel. 2017.
\newblock High performance text recognition using a hybrid convolutional-lstm
  implementation.
\newblock In \emph{2017 14th IAPR international conference on document analysis
  and recognition (ICDAR)}, volume~1, pages 11--16. IEEE.

\bibitem[{Cui et~al.(2019)Cui, Li, Chen, and Zhang}]{cui-etal-2019-fine}
Baiyun Cui, Yingming Li, Ming Chen, and Zhongfei Zhang. 2019.
\newblock \href {https://doi.org/10.18653/v1/D19-1361} {Fine-tune {BERT} with
  sparse self-attention mechanism}.
\newblock In \emph{Proceedings of the 2019 Conference on Empirical Methods in
  Natural Language Processing and the 9th International Joint Conference on
  Natural Language Processing (EMNLP-IJCNLP)}, pages 3548--3553, Hong Kong,
  China. Association for Computational Linguistics.

\bibitem[{Du et~al.(2021)Du, Li, Guo, Cui, Liu, Zhou, Lu, Yang, Liu, Hu, Yu,
  and Ma}]{ppocrv2}
Yuning Du, Chenxia Li, Ruoyu Guo, Cheng Cui, Weiwei Liu, Jun Zhou, Bin Lu,
  Yehua Yang, Qiwen Liu, Xiaoguang Hu, Dianhai Yu, and Yanjun Ma. 2021.
\newblock \href {http://arxiv.org/abs/2109.03144} {Pp-ocrv2: Bag of tricks for
  ultra lightweight {OCR} system}.
\newblock \emph{CoRR}, abs/2109.03144.

\bibitem[{Graves et~al.(2006)Graves, Fern{\'a}ndez, Gomez, and
  Schmidhuber}]{graves2006connectionist}
Alex Graves, Santiago Fern{\'a}ndez, Faustino Gomez, and J{\"u}rgen
  Schmidhuber. 2006.
\newblock Connectionist temporal classification: labelling unsegmented sequence
  data with recurrent neural networks.
\newblock In \emph{Proceedings of the 23rd international conference on Machine
  learning}, pages 369--376.

\bibitem[{He et~al.(2016)He, Zhang, Ren, and Sun}]{he2016resnet}
Kaiming He, Xiangyu Zhang, Shaoqing Ren, and Jian Sun. 2016.
\newblock Deep residual learning for image recognition.
\newblock In \emph{Proceedings of the IEEE conference on computer vision and
  pattern recognition}, pages 770--778.

\bibitem[{Hu et~al.(2021)Hu, Yin, Liu, Li, and Ye}]{hu2021detection}
Chuanbo Hu, Minglei Yin, Bin Liu, Xin Li, and Yanfang Ye. 2021.
\newblock Detection of illicit drug trafficking events on instagram: A deep
  multimodal multilabel learning approach.
\newblock In \emph{Proceedings of the 30th ACM International Conference on
  Information \& Knowledge Management}, pages 3838--3846.

\bibitem[{Kenton and Toutanova(2019)}]{devlin2018bert}
Jacob Devlin Ming-Wei~Chang Kenton and Lee~Kristina Toutanova. 2019.
\newblock Bert: Pre-training of deep bidirectional transformers for language
  understanding.
\newblock In \emph{Proceedings of NAACL-HLT}, pages 4171--4186.

\bibitem[{Li et~al.(2022)Li, Zhou, Li, Li, Liu, Sun, Wang, Li, Cao, and
  Zheng}]{li-etal-2022-past}
Yinghui Li, Qingyu Zhou, Yangning Li, Zhongli Li, Ruiyang Liu, Rongyi Sun,
  Zizhen Wang, Chao Li, Yunbo Cao, and Hai-Tao Zheng. 2022.
\newblock \href {https://doi.org/10.18653/v1/2022.findings-acl.252} {The past
  mistake is the future wisdom: Error-driven contrastive probability
  optimization for {C}hinese spell checking}.
\newblock In \emph{Findings of the Association for Computational Linguistics:
  ACL 2022}, pages 3202--3213, Dublin, Ireland. Association for Computational
  Linguistics.

\bibitem[{Li et~al.(2021)Li, Zhou, Li, Xu, and
  Cao}]{li-etal-2021-improving-bert}
Zhongli Li, Qingyu Zhou, Chao Li, Ke~Xu, and Yunbo Cao. 2021.
\newblock \href {https://doi.org/10.18653/v1/2021.findings-acl.57} {Improving
  {BERT} with syntax-aware local attention}.
\newblock In \emph{Findings of the Association for Computational Linguistics:
  ACL-IJCNLP 2021}, pages 645--653, Online. Association for Computational
  Linguistics.

\bibitem[{Liu et~al.(2020)Liu, Xu, and Zhang}]{liu2020offline}
Brian Liu, Xianchao Xu, and Yu~Zhang. 2020.
\newblock Offline handwritten chinese text recognition with convolutional
  neural networks.
\newblock \emph{arXiv preprint arXiv:2006.15619}.

\bibitem[{Liu et~al.(2011)Liu, Yin, Wang, and Wang}]{liu2011casia}
Cheng-Lin Liu, Fei Yin, Da-Han Wang, and Qiu-Feng Wang. 2011.
\newblock Casia online and offline chinese handwriting databases.
\newblock In \emph{2011 International Conference on Document Analysis and
  Recognition}, pages 37--41. IEEE.

\bibitem[{Loshchilov and Hutter(2018)}]{loshchilov2017adamw}
Ilya Loshchilov and Frank Hutter. 2018.
\newblock Decoupled weight decay regularization.
\newblock In \emph{International Conference on Learning Representations}.

\bibitem[{Lu et~al.(2019)Lu, Batra, Parikh, and Lee}]{lu2019vilbert}
Jiasen Lu, Dhruv Batra, Devi Parikh, and Stefan Lee. 2019.
\newblock Vilbert: Pretraining task-agnostic visiolinguistic representations
  for vision-and-language tasks.
\newblock \emph{Advances in neural information processing systems}, 32.

\bibitem[{Nie et~al.(2021)Nie, Li, Gan, Wang, Zhu, Zeng, Liu, Bansal, and
  Wang}]{nie2021mlp}
Yixin Nie, Linjie Li, Zhe Gan, Shuohang Wang, Chenguang Zhu, Michael Zeng,
  Zicheng Liu, Mohit Bansal, and Lijuan Wang. 2021.
\newblock Mlp architectures for vision-and-language modeling: An empirical
  study.
\newblock \emph{arXiv preprint arXiv:2112.04453}.

\bibitem[{Ratinov and Roth(2009)}]{bio_scheme}
Lev Ratinov and Dan Roth. 2009.
\newblock \href {https://www.aclweb.org/anthology/W09-1119} {Design challenges
  and misconceptions in named entity recognition}.
\newblock In \emph{Proceedings of the Thirteenth Conference on Computational
  Natural Language Learning ({C}o{NLL}-2009)}, pages 147--155, Boulder,
  Colorado. Association for Computational Linguistics.

\bibitem[{Tan and Bansal(2019)}]{tan2019lxmert}
Hao Tan and Mohit Bansal. 2019.
\newblock Lxmert: Learning cross-modality encoder representations from
  transformers.
\newblock In \emph{Proceedings of the 2019 Conference on Empirical Methods in
  Natural Language Processing and the 9th International Joint Conference on
  Natural Language Processing (EMNLP-IJCNLP)}, pages 5100--5111.

\bibitem[{Toto et~al.(2021)Toto, Tlachac, and Rundensteiner}]{toto2021audibert}
Ermal Toto, ML~Tlachac, and Elke~A Rundensteiner. 2021.
\newblock Audibert: A deep transfer learning multimodal classification
  framework for depression screening.
\newblock In \emph{Proceedings of the 30th ACM International Conference on
  Information \& Knowledge Management}, pages 4145--4154.

\bibitem[{Tsai et~al.(2019)Tsai, Bai, Liang, Kolter, Morency, and
  Salakhutdinov}]{tsai2019mmtransformer}
Yao-Hung~Hubert Tsai, Shaojie Bai, Paul~Pu Liang, J~Zico Kolter, Louis-Philippe
  Morency, and Ruslan Salakhutdinov. 2019.
\newblock Multimodal transformer for unaligned multimodal language sequences.
\newblock In \emph{Proceedings of the conference. Association for Computational
  Linguistics. Meeting}, volume 2019, page 6558. NIH Public Access.

\bibitem[{Vaswani et~al.(2017)Vaswani, Shazeer, Parmar, Uszkoreit, Jones,
  Gomez, Kaiser, and Polosukhin}]{vaswani2017attention}
Ashish Vaswani, Noam Shazeer, Niki Parmar, Jakob Uszkoreit, Llion Jones,
  Aidan~N Gomez, {\L}ukasz Kaiser, and Illia Polosukhin. 2017.
\newblock Attention is all you need.
\newblock \emph{Advances in neural information processing systems}, 30.

\bibitem[{Wu et~al.(2013)Wu, Liu, and Lee}]{confusion_set}
Shih-Hung Wu, Chao-Lin Liu, and Lung-Hao Lee. 2013.
\newblock \href {https://aclanthology.org/W13-4406} {{C}hinese spelling check
  evaluation at {SIGHAN} bake-off 2013}.
\newblock In \emph{Proceedings of the Seventh {SIGHAN} Workshop on {C}hinese
  Language Processing}, pages 35--42, Nagoya, Japan. Asian Federation of
  Natural Language Processing.

\bibitem[{Xu et~al.(2021)Xu, Li, Zhou, Li, Wang, Cao, Huang, and Mao}]{csc}
Heng-Da Xu, Zhongli Li, Qingyu Zhou, Chao Li, Zizhen Wang, Yunbo Cao, Heyan
  Huang, and Xian-Ling Mao. 2021.
\newblock \href {https://doi.org/10.18653/v1/2021.findings-acl.64} {Read,
  listen, and see: Leveraging multimodal information helps {C}hinese spell
  checking}.
\newblock In \emph{Findings of the Association for Computational Linguistics:
  ACL-IJCNLP 2021}, pages 716--728, Online. Association for Computational
  Linguistics.

\end{thebibliography}

\clearpage

\appendix

\section{Annotation Process}
We present the process of annotation in our sequence labeling task in Figure \ref{fig:annotate}. We put a label to individual character in textual answer referring to hand-written content and each label stands for a kind of editing operation. 

\cjksong{
\begin{figure*}[!htbp]
    \centering
    \includegraphics[width=0.8\textwidth]{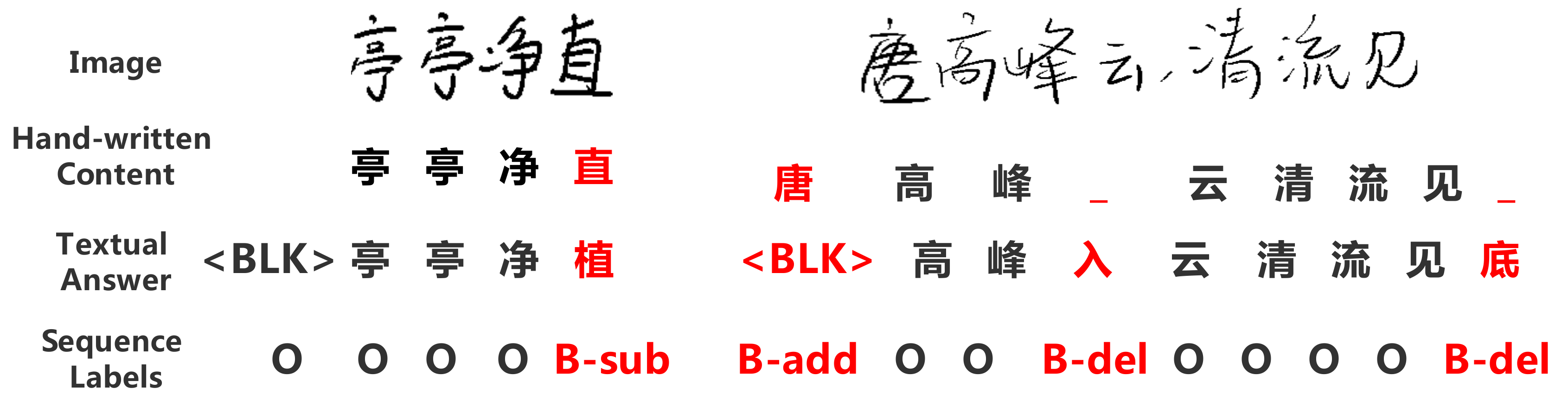}
    \caption{\label{fig:annotate}Examples of sequence labeling annotation. For the first sample, to convert the answer to the hand-written content, the last character in the answer ``植'' should be replaced by ``直''. So the label `B-sub' is annotated to ``植'' and `O' for the rest. For the latter one, the character ``入''  and ``底'' do not appear in the hand-written content so the labels are both 'B-del'. Moreover, the character ``唐'' should be inserted at the first position in answer for conversion, so we put a `B-add' label to the placeholder `<BLK>'.}
\end{figure*}
}

\section{SynCC Data Construction}
\label{appendix-data}
To build a synthetic OCR dataset, we first collect handwritten character images from the HWDB 1.0 set of CASIA-HWDB~\cite{liu2011casia} and our educational application. Then the character images are spliced together into text-line images according to the sentences or clauses of our online essay corpus. We also replace characters with shape-similar ones and their images to enhance OCR models to recognize them. The synthetic OCR examples are shown on Table~\ref{tab:syncc-ocr}. These samples are taken as the positive CCC samples, and negative sample augmentation introduced in Section~\ref{sec:data-aug} is applied to construct SynCC dataset.

\cjksong{
\begin{table}[htp]
    \centering
    \begin{tabular}{ll}
    \toprule
         Image: & \includegraphics[width=26mm,height=5mm]{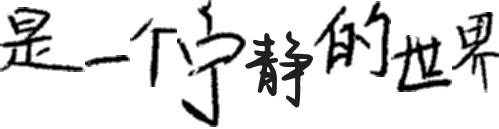} \\
         Content: & 是一个宁静的世界 \\
         Image: & \includegraphics[width=26mm,height=5mm]{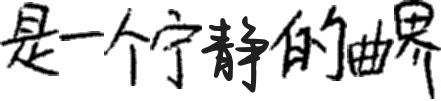} \\
         Content: & 是一个宁静的曲界 \\
         
         \midrule
         
         Image: & \includegraphics[width=24mm,height=5mm]{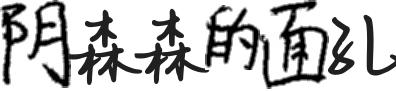} \\
         Content: & 阴森森的面孔  \\
         Image: & \includegraphics[width=24mm,height=5mm]{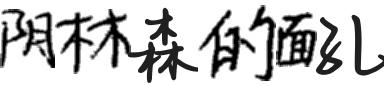} \\
         Content: & 阴林森的面孔  \\
         \midrule         

         Image: & \includegraphics[width=40mm,height=5mm]{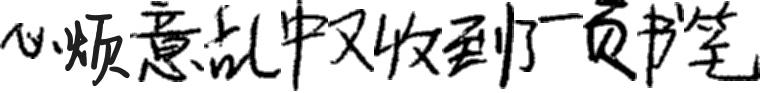} \\
         Content: & 心烦意乱中又收到了一页书笔  \\
         Image: & \includegraphics[width=40mm,height=5mm]{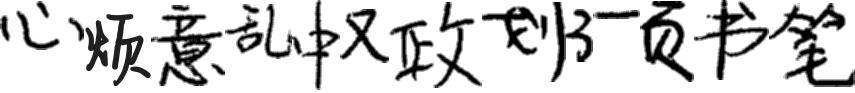} \\
         Content: & 心烦意乱中又政到了一页书笔  \\
         \midrule         

         Image: & \includegraphics[width=35mm,height=5mm]{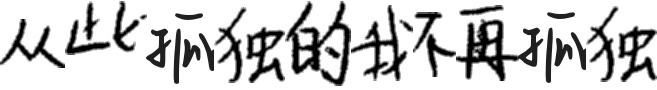} \\
         Content: & 从此孤独的我不再孤独  \\
         Image: & \includegraphics[width=35mm,height=5mm]{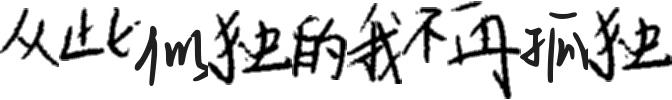} \\
         Content: & 从此似独的我不再孤独  \\

         \bottomrule
    \end{tabular}
    \caption{Synthetic OCR examples of SynCC.}
    \label{tab:syncc-ocr}
\end{table}
}

\section{Recall Drop on EinkCC}
Table~\ref{tab:overall-performance} shows that \model{} improves the correction precision and F1 but the recall drops on EinkCC. We present several false-positive predictions on Table~\ref{tab:fp_einkcc}.
All examples are annotated as wrong by teachers but are predicted as right by \model{}.
In the first example, the student writes down ``\cjksong{斯}'' but there is a extremely large margin between the left and right sides. In the rest examples, students write down shape-similar characters but \model{} can't detect them and tend to predict as right. Thus, even though we use a confusion set to enhance \model{} to detect shape-similar characters, there is still a lot of room for improvement in this regard.

\cjksong{
\begin{table}[htp]
    \centering
    \begin{tabular}{l c c}
    \toprule
         Image: & \includegraphics[width=30mm,height=5mm]{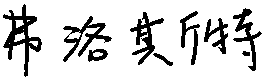} \\
         Answer: & 弗洛{\bf\underline{斯}}特 \\
         Labels:& O~O~O~\underline{B-sub}~O \\
         \midrule
         
         Image: & \includegraphics[width=6mm,height=5mm]{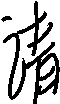} \\
         Answer: & \bf\underline{清} \\
         Labels:& O~\underline{B-sub} \\
         \midrule
         
         Image: & \includegraphics[width=45mm,height=5mm]{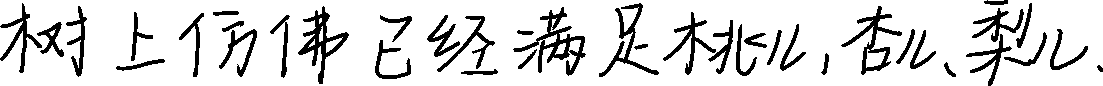} \\
         Answer: & ... 已经满{\bf\underline{是}}桃儿 ... \\
         Labels:& ... O~O~O~\underline{B-sub}~O~O ... \\
         \midrule         
         
         Image: & \includegraphics[width=30mm,height=5mm]{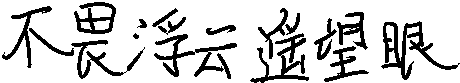} \\
         Answer: & 不畏浮云{\bf\underline{遮}}望眼 \\
         Labels:&  O~O~O~O~O~\underline{B-sub}~O~O \\
         \bottomrule
    \end{tabular}
    \caption{False-positive predictions on EinkCC test set. All above answer characters are predicted to `O' by \model{}. The character marked by the underline indicates why the correction result should be wrong.}
    \label{tab:fp_einkcc}
\end{table}
}

\section{Predictions of \model{}}
Table \ref{tab:sequence-case} shows several cases of \model{}'s predictions. \model{} is able to identify all kinds of modifications between the answer and hand-written content including substitutions, deletions and insertions, which proves that it's reasonable to format the fill-in-the-blank assignments correction task to sequence labeling. In this way, \model{} can not only indicate the correctness of students' answers, but also can locate where the errors occur.

\cjksong{
\begin{table*}[htp]
    \centering
    \begin{tabular}{c c c}
    \toprule
         & Image: & \includegraphics[width=30mm,height=5mm]{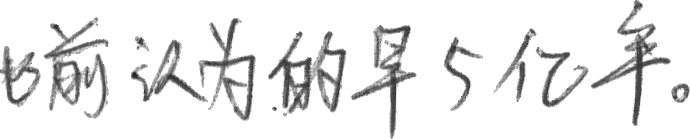} \\
         &Handwritten Content: & 前~认~为~的~早~5~亿~年  \\
         &Correct Answer: & {\bf\underline{麻}}~前~{\bf\underline{镅}}~认~为~的~早~5~亿~{\bf\underline{材}}~{\bf\underline{嚓}}~年\\
         &\model{} Labels:& O B-del O B-del I-del O O O O O O B-del I-del O\\
         &\model{} Predictions:& O B-del O B-del I-del O O O O O O B-del I-del O\\
         \midrule
         
         & Image: & \includegraphics[width=15mm,height=5mm]{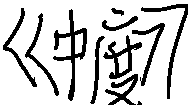} \\
         &Handwritten Content: &《~中~度~》 \\
         &Correct Answer: & 《~中~{\bf\underline{庸}}~》\\
         &\model{} Labels:& O O O B-sub O\\
         &\model{} Predictions:& O O O B-sub O\\
         \midrule
         
         & Image: & \includegraphics[width=23mm,height=5mm]{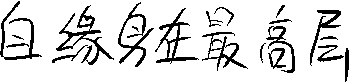} \\
         &Handwritten Content: &自~缘~身~在~最~高~层~\\
         &Correct Answer: & {\bf\underline{不~畏~浮~云~遮~望~眼~}}\\
         &\model{} Labels:& O B-sub I-sub I-sub I-sub I-sub I-sub I-sub\\
         &\model{} Predictions:& O B-sub I-sub I-sub I-sub I-sub I-sub I-sub\\
         \midrule         
         
         & Image: & \includegraphics[width=18mm,height=5mm]{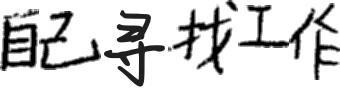} \\
         &Handwritten Content: &自~己~寻~找~工~作~\\
         &Correct Answer: &{\bf\underline{自~寻}}~工~作~\\
         &\model{} Labels:& O B-add B-add O O\\
         &\model{} Predictions:& O B-add B-add O O\\
         \bottomrule
    \end{tabular}
    \caption{Several outputs of \model{} on dev and test sets. Some answers are generated through our data augmentation. \model{} can identify editing operations between answers and hand-written content. Characters are shown in bold with underline if the corresponding label is not `O'.}
    \label{tab:sequence-case}
\end{table*}
}

\end{document}